\documentclass[10pt,twocolumn,letterpaper]{article}

\usepackage{cvpr}
\usepackage{times}
\usepackage{epsfig}
\usepackage{graphicx}
\usepackage{amsmath}
\usepackage{amssymb}

\usepackage[pagebackref=true,breaklinks=true,letterpaper=true,colorlinks,bookmarks=false]{hyperref}

\cvprfinalcopy 

\begin{document}

\title{Interpretable Partitioned Embedding for Customized Fashion Outfit Composition}

\author{$^{\dag}$Zunlei Feng, $^{\dag}$Zhenyun Yu, $^{\ddag}$Yezhou Yang, $^{\dag}$Yongcheng Jing, $^{\flat}$Junxiao Jiang, $^{\dag}$Mingli Song\\
$^{\dag}$Zhejiang University, Hangzhou, China\\
$^{\ddag}$Arizona State University, Phoenix, Arizona\\
$^{\flat}$Alibaba Group, Hangzhou, China
}

\maketitle

\begin{abstract}
Intelligent fashion outfit composition becomes more and more popular in these years. Some deep learning based approaches reveal competitive composition recently. However, the unexplainable characteristic makes such deep learning based approach cannot meet the the designer, businesses and consumers' urge to comprehend the importance of different attributes in an outfit composition. To realize interpretable and customized fashion outfit compositions, we propose a partitioned embedding network to learn interpretable representations from clothing items. The overall network architecture consists of three components: an auto-encoder module, a supervised attributes  module and a multi-independent  module. The auto-encoder  module serves to encode all useful information into the embedding. In the supervised attributes module, multiple attributes labels are adopted to ensure that different parts of the overall embedding correspond to different attributes. In the multi-independent module, adversarial operation are adopted to fulfill the mutually independent constraint. With the interpretable and partitioned embedding, we then construct an outfit composition graph and an attribute matching map. Given specified attributes description, our model can recommend a ranked list of outfit composition with interpretable matching scores. Extensive experiments demonstrate that 1) the partitioned embedding have unmingled parts which corresponding to different attributes and 2) outfits recommended by our model are more desirable in comparison with the existing methods.
\end{abstract}

\section{Introduction}

\emph{``Costly thy habit as thy purse can buy, But not expressed in fancy; rich, not gaudy, For the apparel oft proclaims the man.''}

\rightline{William Shakespeare}

 Fashion style tells a lot about one's personality. With the fashion industries going online, clothing fashions are becoming a much more popular topic among the general public. 
There have been a number of research studies on clothes retrieval and recommendation~\cite{di2013style,fu2012efficient,hadi2015buy,huang2015cross,liu2017deepstyle,wang2011clothes}, clothing category classification~\cite{chen2012describing,kiapour2014hipster,liu2016deepfashion,simo2016fashion,yamaguchi2012parsing}, attribute prediction ~\cite{abe2017changing,bossard2012apparel,chen2015deep,yamaguchi2015mix} and clothing fashion analysis~\cite{liu2017deepstyle,matzen2017streetstyle}. However, due to the fact that the fashion concept is often subtle and subjective, the composition of fashion outfit keeps being an open problem to reach consensus for the general public.
%
%
%
%
%
%

%
%
%
%
 \begin{figure}
\centering
\includegraphics[scale =0.32]{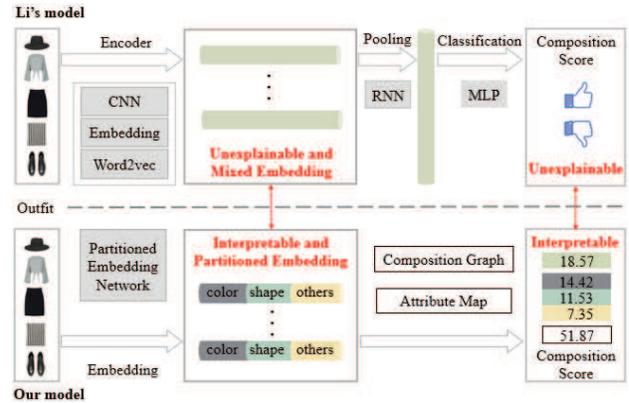}
\caption{Comparative framework between our model and Li's ~\cite{li2017mining} model. Features used by Li's method are usually mixed and unexplainable, which lead to unexplainable outfit composition. Different parts of embedding extracted by our model correspond to different attributes. With the partitioned embedding, our model can recommend outfit composition with interpretable matching scores.}
\label{fig:compareArch}
\end{figure}

Until now, there are few works ~\cite{iwata2011fashion,li2017mining} studying fashion outfit composition. Iwata \emph{et al}. ~\cite{iwata2011fashion} present an approach by concatenating hand-crafted features into a vector as an embedding for each clothing item.  The extracted specified attribute features of this approach are usually mixed with other attribute features. Very recently, Li \emph{et al}.~\cite{li2017mining} present a deep neural network based method by adopting mixed multi-model embedding to represent an outfit item as a whole. There are two characteristics  for such deep neural network based embedding method: 1) the embedding is unexplainable; 2) the embedding does not bring attribute information.  Unfortunately, in many practical applications, it is necessary to understand the importance of different attributes in an outfit composition for designers, businesses and consumers. That is to say, an interpretable and partitioned embedding is vital for a practical fashion outfit composition system.
%
%
%
%

To address the aforementioned problems, we proposed a partitioned embedding network in this paper. The proposed network architecture consists of three components: an auto-encoder module, a supervised attributes module, and a multi-independent module. The auto-encoder module serves to encode all useful information into the embedding. In the supervised attributes module, multiple attributes labels are adopted to ensure that different parts of the overall embedding correspond to different attributes. In the multi-independent module, to ensure each part of the embedding only relates to the corresponding attribute, the mutually independent constraint is taken into account. Then, considering that matching items may appear multiple times in different outfits, we propose a fashion composition graph to model matching relationships in outfit with the extracted partitioned embeddings. Meanwhile, an attribute matching map which learns the importance of different attributes in the composition is also built. Comparative framework between our model and the multi-modal embeddings based method ~\cite{li2017mining} is shown in Figure ~\ref{fig:compareArch}.
%
%
%
%

To summarize, our work has three primary contributions:
1) We present a partitioned embedding network, which can extract interpretable embeddings of fashion outfit items.
2) We put forward a weakly-supervised fashion outfit composition model, which depends solely on a large number of outfits without quality scores of outfits as others. Besides, our model can be extended to a dataset with annotated quality scores.
%
%
%
%
3) An iterative and customized fashion outfit composition scheme is given. Since fashion trends alter directions quickly and dramatically, our model can keep up with the fashion trends by easily incorporating new fashion outfit dataset. 
%
%
%
%

\section{Related Work}

%
%
%
%

{\bf Embedding Methods.}
In recent years, there are several models~\cite{matzen2017streetstyle,simo2016fashion,vittayakorn2015runway} can be used to get embeddings of clothes. Vittayakorn \emph{et al}. ~\cite{vittayakorn2015runway} extract five basic feature (color, texture, shape, parse and style) of outfit appearance and concatenated them to form a vector for representing outfit.  They use the vector to learning the outfit similarity and analyze visual trends in fashion. Matzen \emph{et al}. ~\cite{matzen2017streetstyle} adopt deep learning method to train several attribute classifiers and used high-level features of the trained classifiers to create a visual embedding of clothing style. Then, using the embedding, millions of Instagram photos of people sampled worldwide are analyze to study spatio-temporal trends in clothing around the globe. Simo-Serra \emph{et al}. ~\cite{simo2016fashion} train a classifier network with ranking as the constraint to extract discriminative feature representation and also used high level feature of the classifier network as embeddings of fashion style. Other embedding methods that are related to our work include Auto-Encoder ~\cite{bengio2009learning}, Variational Auto-Encoder (VAE) ~\cite{kingma2013auto} and Predictability Minimization (PM) model ~\cite{schmidhuber2008learning}. Autoencoder and VAE are used to encode embeddings from unlabeled data. The encoded embeddings usually contain mixed and unexplainable features of original images. To get partitioned embeddings, Schmidhuber ~\cite{schmidhuber2008learning} adopted an adversarial operation to get embeddings, where units are independent. However, the independent units are not meaningful.

{\bf Fashion Outfit Composition.}
As described above, due to the difficulty in modeling outfit composition, there are few works ~\cite{iwata2011fashion,li2017mining,simo2015neuroaesthetics}  studying fashion outfit composition. Iwata \emph{et al}.~\cite{iwata2011fashion} propose a topic model to recommend "Tops" for "Bottoms". The goal of this work is to compose fashion outfit automatically. challenging in modeling many aspects of the fashion outfits, such as compatibility and aesthetics. Veit \emph{et al}.~\cite{veit2015learning} use a Siamese Convolutional Neural Network (CNN) architecture to learn clothing matching from the Amazon co-purchase dataset, focusing on the representative problem of learning compatible clothing style. Simo-Serra1 \emph{et al}.~\cite{simo2015neuroaesthetics} introduce a Conditional Random Field (CRF) to learn the different outfits, types of people and settings. The model is used to predict how fashionable a person looks on a particular photograph. Li \emph{et al}.~\cite{li2017mining} use the quality scores as the label and multi-modal embeddings as features to train a grading model. The framework of Li's approach is shown in Figure ~\ref{fig:compareArch}(a). However, the mixed multi model embedding used by Li's model is unexplainable, which leads to unexplainable outfit composition.

%
%
%
%

\begin{figure*}
\centering
\includegraphics[scale =0.51]{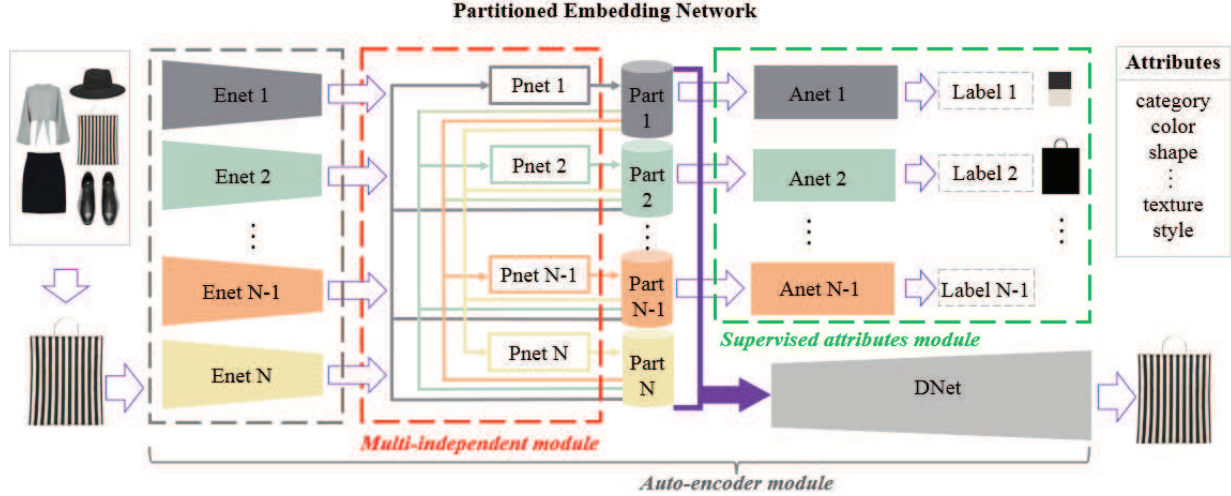}
\caption{The framework of partitioned embedding network. The overall network architecture consists of three components: an auto-encoder module, a supervised attributes module and a multi-independent module. The auto-encoder module serves to encode all useful information into the embedding. It is composed of encoder networks $Enet\ 1, Enet\ 2,..., Enet\ N$ and decoder network $DNet$. In the supervised attributes module, attributes networks $Anet\ 1, Anet\ 2,..., Anet\ N-1$ and labels $Label\ 1, Label\ 2,..., Label\ N-1$ are used to ensure that different parts of the overall embedding correspond to different attributes. In the multi-independent block, adversarial prediction networks $Pnet\ 1, Pnet\ 2,..., Pnet\ N$ are adopted to make sure that different parts of whole embedding are independent.}
\label{fig:Framework}
\end{figure*}

\section{The Proposed Method}
To overcome the limitations of existing fashion outfit composition approaches, we propose an interpretable partitioned embedding method to achieve customized fashion outfit composition.
In section 3.1, the partitioned embedding network is firstly presented how to partition an embedding into  interpretable  parts which correspond to different attributes. Then, we introduce how to build composition relationship in our proposed composition graph and attributes maps with the interpretable and partitioned embedding.
%
%
%
%
%

\subsection{Interpretable Partitioned Embedding}
Let $I$ denote an item of outfit, $R(I)$ denote the encoded embedding of item $I$.  To make the extracted embedding interpretable and partitionable, there are two constraints that should be satisfied: on one hand, the fixed parts of whole embedding should correspond to specific attributes; on the other hand, different parts of the whole embedding should be mutual independent. Thus, the embedded attributes embedding $R(I)$ of item $I$ can be described as below:
%
%
%
%
\begin{eqnarray}
R(I) =[r_1;r_2;...;r_N],\\ \label{eq_6}
s.t. P(r_{i}| \{ r_{j}, j \neq i\})=P(r_{i})\nonumber
\end{eqnarray}
where $r_{i}$ corresponds to different parts of embedding and $N$ is the total number of attributes. Condition $P(r_{i}| \{ r_{j}, j \neq i\})=P(r_{i})$ implies that $r_{i}$ does not depend on $\{ r_{j}, j \neq i\}$. Figure ~\ref{fig:Framework} shows framework of the interpretable partitioned embedding network. The whole embedding network is an auto-encoder network which embeds items of outfits into another feature space. The whole encoder network is composed of attribute encoder networks $Enet\ 1, Enet\ 2,..., Enet\ N$, which embeds an original item into different parts $r_{k},\ k \in \{1,2,...,N\}$ of whole embedding. The decoder network $DNet$ decodes an whole embedding back to original image. In the process of decoding, attributes networks $Anet\ 1, Anet\ 2,..., Anet\ N-1$ and labels $Label\ 1, Label\ 2,..., Label\ N-1$ serve to learn useful features which are related to corresponding attributes. Meanwhile, the mutually independent constraint is taken into account, which can not only ensure different parts of embedding solely related to corresponding attribute but also extract embedding of indefinite attributes, such as, texture and style. As a consequence, the loss of the partitioned embedding network is defined as follows:
%
%
%
%
\begin{equation}
L=L_{i}(I,I')+\alpha L_{t}(I,Label)+\beta L_{d}(r_1,r_2,...,r_N), \label{eq_4}
\end{equation}
where $\alpha$, $\beta$ are balancing parameters, $I'$ is the decoded image of item $I$, $L_{i}(I,I')$ is the auto-encoder loss, $L_{t}(I,Label)=\frac{ \sum_{k=1}^{N}loss(I, Label\ k)}N$ is the summed loss of each auxiliary attributes loss(I, Label\ k) and $L_{d}(r_1,r_2,...,r_N)$ is mutually independent loss. Inspired by the adversarial approach in ~\cite{schmidhuber2008learning}, we adopt adversarial operation to meet the mutually independent constraint. The mutually independent loss $L_{d}(r_1,r_2,...,r_N)$ includes predicting loss $L_{p}$ and encoding loss $L_{e}$. In the prediction stage, each prediction net $Pnet\ i$ tries to predict corresponding embedding part $r_i$ as much as possible to maximize predictability. The predicted loss $L_{p}$ can be defined as:
%
%
%
%
\begin{equation}
L_{p}=\sum_{i=1}^{N} \mathbb{E}[r_i-f_{i}(r_1,...,r_{i-1},r_{i+1},...,r_{N})]^2, \label{eq_4}
\end{equation}
where $f_{i}$ is function representation of prediction net $Pnet\ i$. In the encoding stage, all encoder nets $Enet i$ try to make all prediction net $f_i$ fail to predict corresponding embedding, which means to minimize predictability. The encoded loss $L_{e}$ thus could be defined as:
\begin{equation}
L_{e}=-\sum_{i=1}^{N}  \mathbb{E}[r_i-f_{i}(r_1,...,r_{i-1},r_{i+1},...,r_{N})]^2. \label{eq_4}
\end{equation}

In summary, the auto-encoded loss $L_{i}(I,I')$ makes sure that the whole embedding contains all information in the item of an outfit. Attributes sum loss $L_{t}(I,Label)$ serves to learn useful information which are related to corresponding attributes. Mutually independent loss $L_{d}(r_1,r_2,...,r_N)$ ensures that different parts of embedding only related to corresponding attributes.  

\subsubsection{niuniuniu}

As aforementioned, there is a large number of attributes~\cite{vittayakorn2015runway,liu2016deepfashion,matzen2017streetstyle} for describing fashion, such as category, color, shape, texture, style and so on. Considering that some attributes are indefinable (such as texture and style), we classify those attributes into same class as remaining attributes. The information of remaining attributes can be extracted with the mutually independent constraint. To get importance of attributes in the fashion cloth, we administer a questionnaire among 30 professional stylists. According to results of the questionnaire, we split attributes into 4 classes and rank them as follow: category, color, shape, texture and remaining attributes (texture, style and others). Thus, in the experiment, we split the whole dataset into different groups according to the category. So, the number of attribute $N$ equals to 3. $Enet\ 1, Enet\ 2, Enet\ 3$ are encoding networks of the color attribute, shape attribute and remaining attributes( texture, style and so on), respectively. $Anet\ 1, Anet\ 2, Anet\ 3$ are supervised networks of the color attribute, shape attribute and remaining attributes, respectively. $PNet\ 1, PNet\ 2, PNet\ 3$ are partitioning network of color attribute, shape attribute and remaining attributes, respectively.
%
%
%
%
%
%
%
%

\begin{figure}
\centering
\includegraphics[scale =0.29]{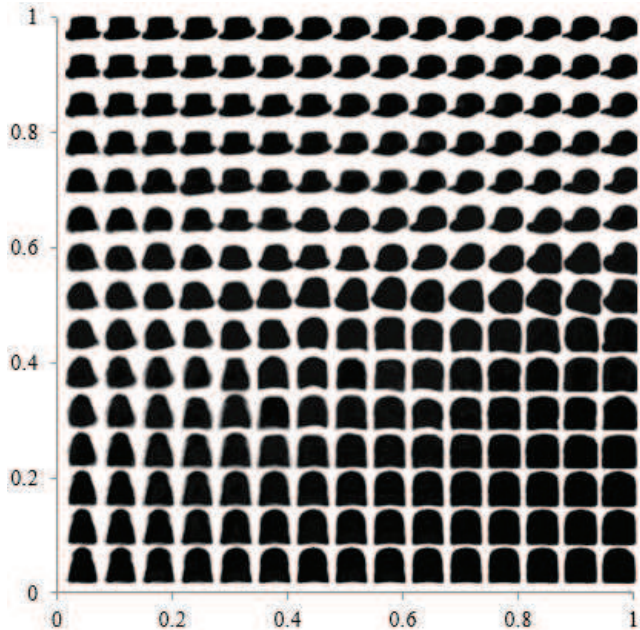}
\caption{Encoded shapes of hat by VAE model with uniform sampling of latent embedding.}
\label{fig:autoEncoderHats}
\end{figure}

{ \bf Color embedding extraction.} As described above, color is a primary element in fashion outfit composition. To get color label, we adopt latest proposed color theme extraction method~\cite{Feng2017Finding} to extract color themes. This method can extract large span and any number of ranked color themes. We modify it to just extract color in the foreground area of the item. Meanwhile, we adopt top-5 extracted color themes of an item as the color label. In the experiment, we adopt Generative adversarial nets(GAN)~\cite{goodfellow2014generative} to extract color embedding. So, the color attribute supervised network $Anet\ 1$ is a GAN architecture, which includes a generative color model $G_{c}$ and a color discriminative model $D_{c}$. Input and output of network $G_{c}$ are $Part\ 1$ and corresponding color themes, respectively. Input and output of network $D_{c}$ are color themes and a discriminant value, respectively. The architectures of these two networks are summarized in Table 1.
%
%
%
%
%
%
%
%

\begin{figure}
\centering
\includegraphics[scale =0.40]{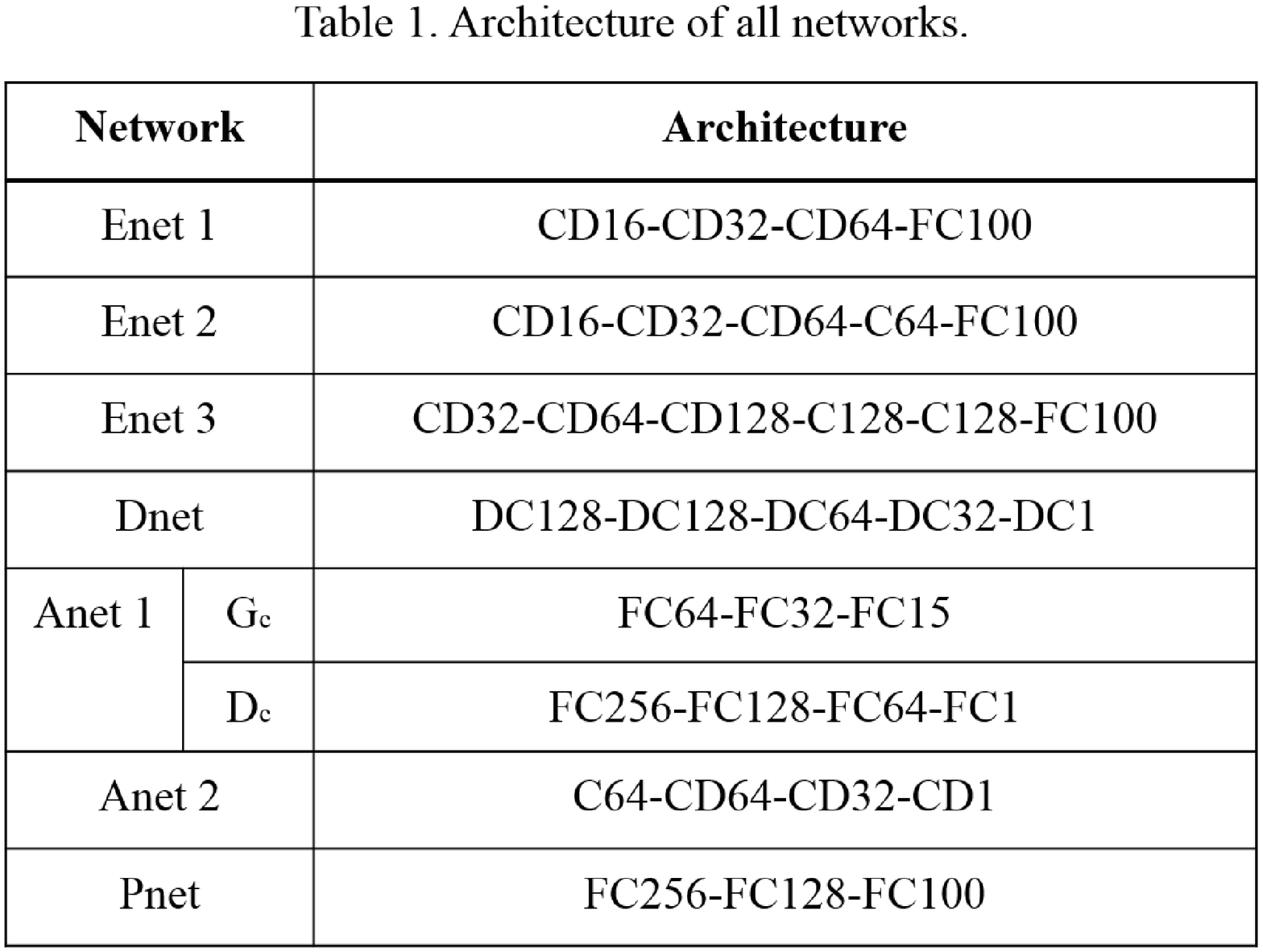}
\label{fig:networkArch}
\end{figure}

{ \bf Shape embedding extraction.} To get an explicit mapping relationship between shape and embedded codes, we adopt a Variational Auto-Encoder model ~\cite{kingma2013auto} to encode and decode shape information. We conduct a toy experiment to validate that shape network can encode all shapes in shape space. The toy VAE network encode the original hats items into latent embeddings with two parameters. Then, through uniform sampling of two parameters, corresponding encoded shapes are shown in  Figure ~\ref{fig:autoEncoderHats}, where we can see that almost all kinds of hats' shape are included. Meanwhile, shapes information are encoded and decoded relatively satisfactory. For each item of an outfit, we use the basic open-close operation and threshold methods to get the mask of item as the label of shape. Input and output of the shape attribute network $Anet\ 2$ are $Part\ 2$ and corresponding shape mask, respectively.
%
%
%
%

{ \bf Remaining attributes embedding extraction.}
As described above, indefinable attributes (texture, shape and others) are classify into same class as remaining attributes. To extract corresponding information of remaining attributes, mutually independent constraint is took into account. In this article, adversarial method is adopted to realize mutually independent constraint. For each attribute corresponding embedding $Pnet\ i$, there is a prediction network $Pnet \ i$ that ensure $Part\ i$ is independent of all other $\{Part\ j, j \neq i\}$. All the prediction networks share the same architecture. Input and output of prediction network $Pnet \ i$ are $\{Part\ j, j \neq i\}$, $Part\ i$.
%
%
%
%

Let $Ci$ denote a Convolution-BatchNorm-ReLU layer with $i$ filters. $CDj$ denotes a Convolution-BatchNorm-Dropout-ReLU layer with a dropout rate of 30\% and $j$ filters. $DCk$ denotes a deconvolution-ReLU layer with $k$ filters, and $FCt$ denotes a Full Connection with $t$ neuron. All convolutions are 4x4 spatial filters with stride 2. Convolutions in both encoder and the discriminator are downsampled by a factor of 2, whereas in the decoder they upsample by a factor of 2. Input image size of $Enet\ 1, Enet\ 2 and Enet\ 3$ is 128*128*3. All the network architectures are summarized in table 1.
%
%
%
%
%

\subsection{Fashion Outfit Composition}

Considering that some matching items may appear many times in different outfits, we propose a fashion composition graph to model matching relationship in outfit. For each category, we first cluster all items into ${N_{p}}$ cluster centers $P_{k}, k\in \{1,2,3,...,N_{p}\}$ according to color embedding. Then, for each cluster $P_{k}$, items that belongs to it is clustered into $N_{q}$ cluster centers $Q_{k}, k\in \{1,2,3,...,N_{q}\}$ according to shape embedding. Lastly, items belonging to shape cluster center are clustered into $N_{r}$ cluster centers $R_{k}, k\in \{1,2,3,...,N_{r}\}$ according to the remaining embedding. Then, we can get $\mathcal{N}=N_{p}N_{q}N_{r}$ cluster centers $C_{k}, k\in \{1,2,3,...,\mathcal{N}\}$. After getting all clustering centers, fashion composition graph $G$ is defined as follow:
\begin{equation}
G=(V,E,W),
\end{equation}
where $V$ is the vertex set of all cluster center $C_{k}$, $E$ is the edge set and $W$ is the weight set.

\begin{figure}
\centering
\includegraphics[scale =0.27]{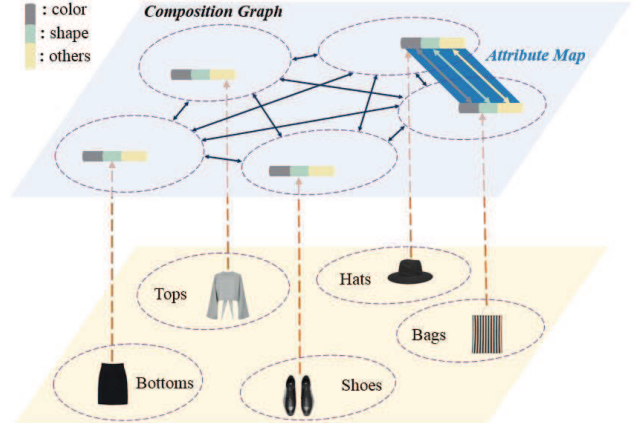}
\caption{Fashion outfit composition graph. In the graph, all items are classified into five classes according to the category. For each category, items are clustered into different cluster centers according to attributes' importance. With these cluster centers as vertexes, edges and weights of fashion composition graph are learned from outfit dataset. Meanwhile, attributes maps between items are built, which model significance of different attributes.}
\label{fig:compositionGraph}
\end{figure}

At the initial stage, all vertexes $v\in V$ have no  connection and all weights $w \in W$ are equal to zeros. If item $I \in v_{i}$ and item $I' \in v_{j}$ appear in the same outfit and there is no connection between $v_{i}$ and $v_{j}$, a connection is created and weight $w_{i,j}$ is set as one. If corresponding vertices $v_{i}$ and $v_{j}$ already have connection, the weight between them will be updated as follow:
\begin{equation}
w_{i,j}=w'_{i,j}+\alpha,
\label{eq:w}
\end{equation}
where $w'_{i,j}$ is the weight in the last stage. Figure ~\ref{fig:compositionGraph} shows an example of the connection process.

{ \bf Attribute Matching Map.}
After obtaining the interpretable and partitioned embeddings, for each category, every attribute class corresponding parts of embeddings are clustered into several clusters. In the process of building fashion outfit graph $G$, an attribute composition score map $M$ of different attributes is also built.  which is defined as:

\begin{eqnarray}
M=\{\{U_{x}^c,A\},\{H_{y}^c,B\},\{O_{z}^c,D\}\}, \\  \label{eq_2}
c \in \{1,2,..,C\},    x \in \{1,2,..,X\}, \nonumber  \\
 y \in \{1,2,..,Y\},   z \in \{1,2,..,Z\}  \nonumber
\end{eqnarray}
where, $\mathcal{C}$ is the number of categories, $X,Y,Z $ are clusters' number of color feature, shape feature and remaining features respectively, $U_{x}^c$ is the $x$-th color attribute cluster of category $c$, $U_{y}^c$ is the $y$-th shape attribute cluster of category $c$, $O_{z}^c$ is the $z$-th other attribute cluster of category $c$, $A, B, D$ are score values set and all initial score value in set are equal to zero.
%
%
%
%
When an item $I$ of category $c$ and an item $I'$ of category $c'$ appear together in the same outfit, score value $a(i,j)$ between color cluster $U_{i}^{c}$ ( $\ r_{1} \in U_{i}^{c},\ R(I)=[r_{1};r_{2};r_{3}]$) and color cluster $U_{j}^{c'}$ ( $r'_{1} \in U_{j}^{c'},\ R(I')=[r'_{1};r'_{2};r'_{3}]$ ) will be updated using the following equation:
\begin{eqnarray}
a(i,j)=a'(i,j)+1, a(i,j)\in A ,
\end{eqnarray}
where,$a'(i,j)$ is last stage score value. $b(i,j)\in B$ and $d\in D$ are updated in the same way.

\begin{figure*}
\centering
\includegraphics[scale =0.59]{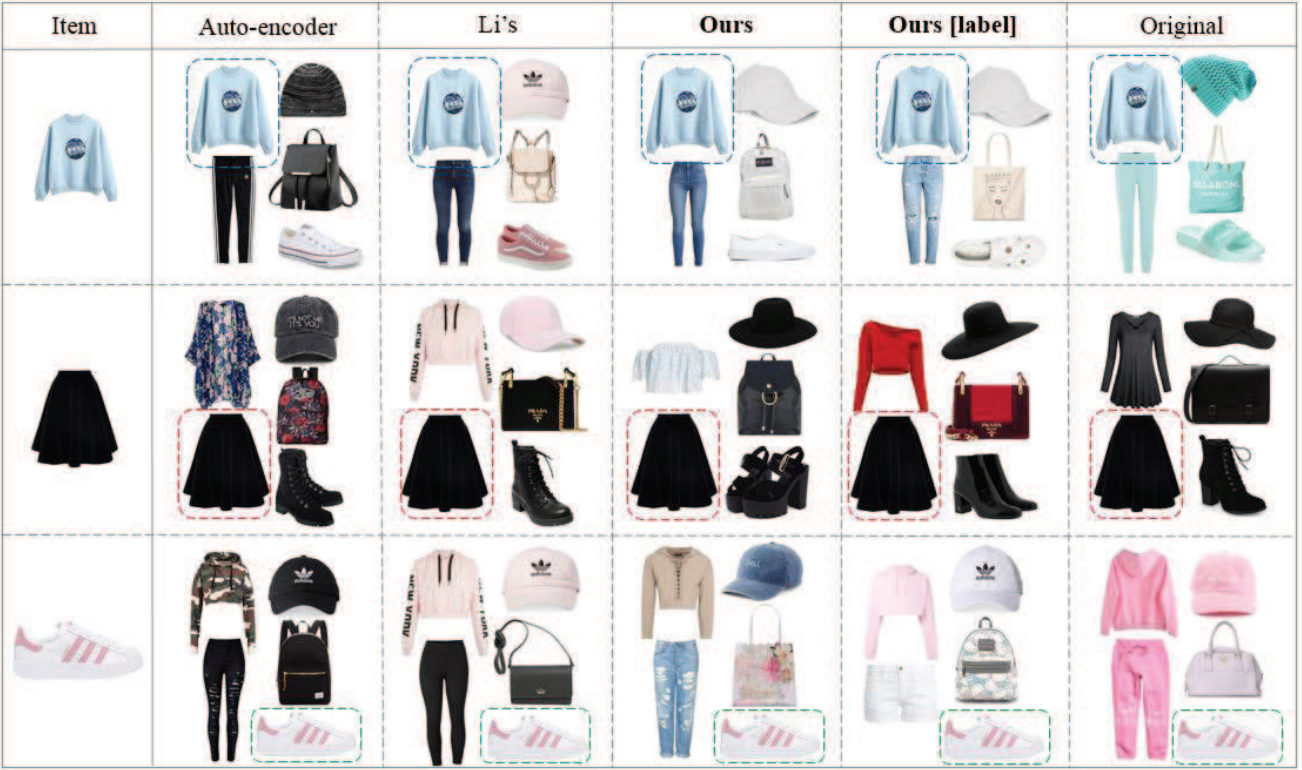}
\caption{Composition visual comparison among different methods}
\label{fig:compositionCompared}
\end{figure*}

After getting outfit composition graph $G$ and attributes map $M$, composition score $S$ of an outfit $u$ is defined as follow:
\begin{eqnarray}
S(u)=S_{1}+\alpha S_{2},  \\
S_{1}=\frac{\sum_{k=1}^{K} w_{k} }{K \sigma(w_1,w_2,...,w_{K})},  \nonumber  \\
S_{2}=\sum_{t=1}^3 \frac{\sum_{k=1}^{K} s^t_{k} }{K \sigma(s^t_1,s^t_2,...,s^t_{K})}  \nonumber
\end{eqnarray}
where, $S_{1}$ is the matching score $S_{1}$ in composition graph, $S_{2}$ is attributes matching scores $S_{2}$. $K$ is the number of matching weights in outfit $o$,  $var(w_1,w_2,...,w_{K})$ is variance among matching weights $w_k, k\in \{1,2,...,K\}$. To get an matching outfit for an item $I$, an exhaustion algorithm is adopted to search the most matching outfit. In order to reduce time complexity, only top $N_{t}$ connected items of each category are taken into account.

\section{Experiments}

\subsection{Implementation Details}

{ \bf DataSet.}
 In the experiment, We collect a dataset from Polyvore.com, which is the most popular fashion oriented website in the US with tens of thousands of fashion outfits creating every day. In the dataset, fashion outfits are associated with the number of $\emph{likes}$  and some fashion items. Each fashion item has corresponding image, category and $\emph{like}$. In practical application, an outfit may contain many items, such as tops (blouse, cardigan, sweater, sweatshirt, hoodie, tank, tunic ), bottoms(pant, skirt, short), hats, bags, shoes, glasses, watches, jewelry and so on. In this article, we choose five prerequisite categories( tops, bottoms, hats, bags and shoes). We perform some simple filtering over raw datasets through discarding items that mix human body. Finally, we get a set of 1.56 million outfits and 7.53 million items.
%
%
%
%
%
%
%
%

{ \bf Setting.}
In our experiment, $\alpha$ is 2 and $\beta$ is 0.7, number of cluster center for color attribute is 1000, number of cluster center for shape attribute is $\sqrt{N_{color\ i}}$, $N_{color\ i}$ is total number of items in cluster $i$, cluster center for remaining attributes is $\sqrt{N_{shape\ j}}$, $N_{shape\ j}$ is total number of items in cluster $j$.

\subsection{Outfit Composition's Psycho-Visual Tests}

To verify the validity of our proposed method, we make a pairwise comparison study. In this experiment, a total of one hundred items are used, along with 20 items for each category. For each item, each method will give an outfit. 30 professional stylists take part in the psychophysical experiment, 13 males and 17 females. In the paired comparison, a stylist is presented with a pair of the recommended outfit. The stylists are asked to choose the most matching composition. All pairs are presented twice to avoid mistakes in individual preferences. The number of pairs is $n(n-1)N_{u}/2$, where $n$ is the number of composition methods and $N_{u}$ is the total number of test outfits. For each pair, the outfit chosen by an observer is given a score of 1, the other outfit is given a score of 0 and both of them got a score of 0.5 when equally matching. After obtaining the raw data, we converted it to frequency matrix $\mathcal{F}=(f_{ij})_{1\leqslant i,j\leqslant n}$, where the entry $f_{ij}$ denotes the score given to composition method $j$ as compared with composition method $i$. From matrix $\mathcal{F}$, percentage matrix $\mathcal{M}=(m_{ij})_{1\leqslant i,j\leqslant n}$ is calculated through $\mathcal{M}=\mathcal{F}/N_{o}$, where $N_{o}$ denotes total number of observations. Table $2$ shows the matrix $\mathcal{M}$ obtained in our experiment.
\begin{figure}
\centering
\includegraphics[scale =0.38]{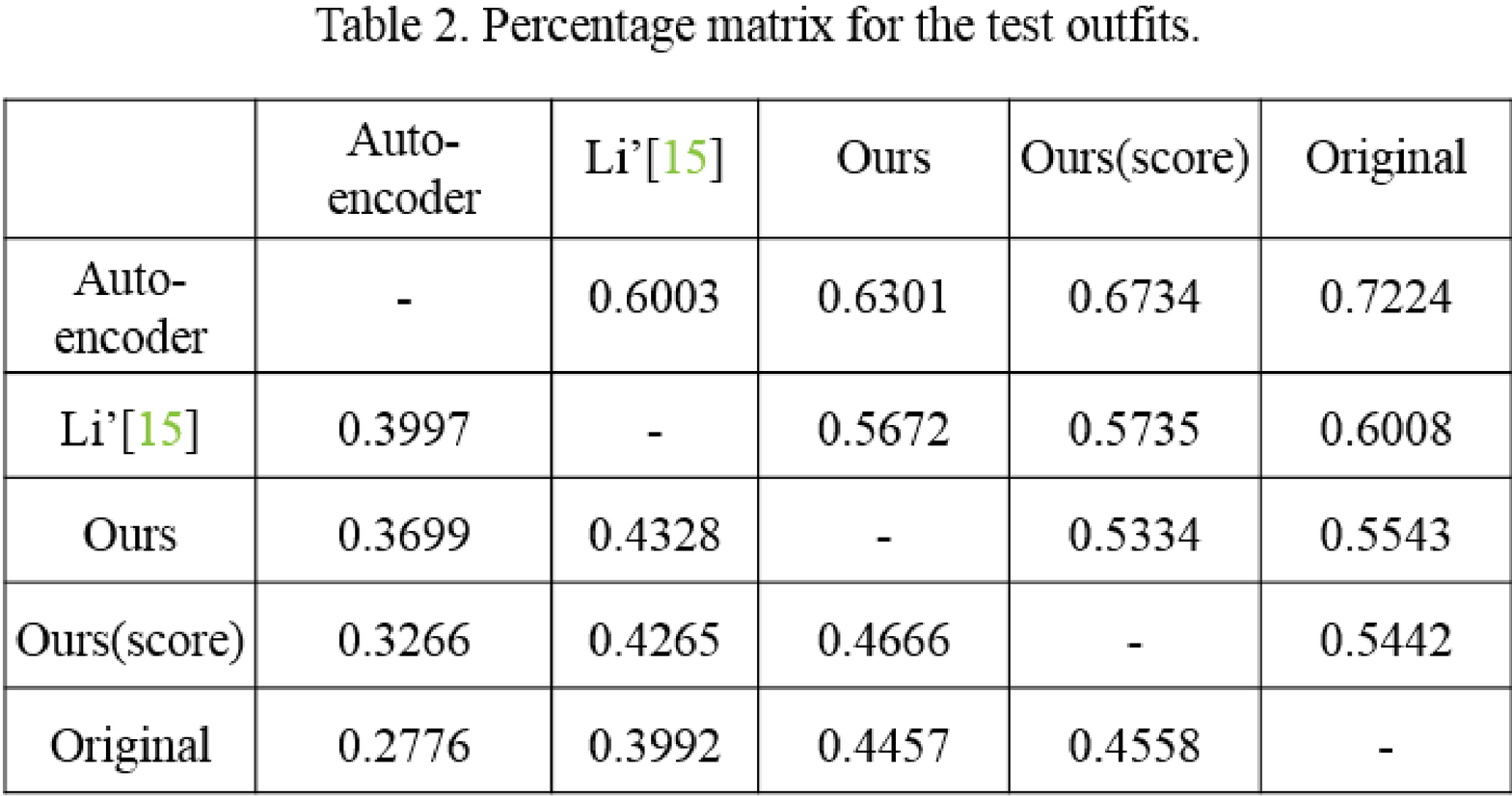}
\label{fig:PM}
\end{figure}
\begin{figure}
\centering
\includegraphics[scale =0.65]{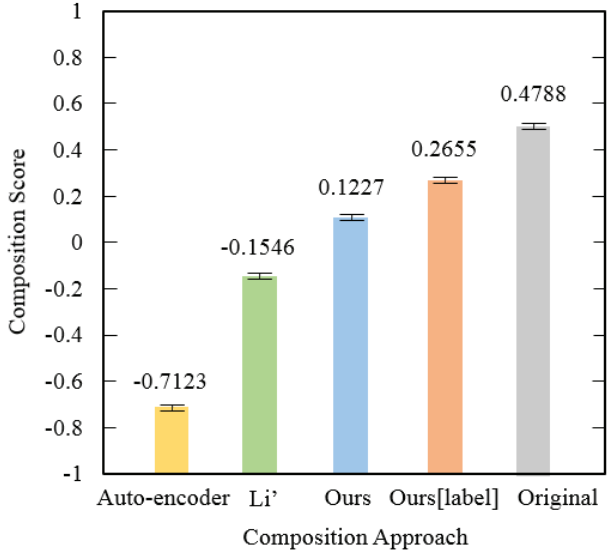}
\caption{Score of composition for the ranking experiment with all professional stylists}
\label{fig:scoreHigtogram}
\end{figure}
To get preference scales from the matrix $\mathcal{M}$, we apply the case V of Thurstone's law of comparative judgment~\cite{thurstone1927law} following Morovic's thesis~\cite{morovic1998develop}. Firstly, a logistic matrix $\mathcal{G}=(g_{ij})_{1\leqslant i,j\leqslant n}$ is calculated through Eq.(\ref{eq_g}) as
\begin{equation}
g_{ij}=ln\left(\frac{N_{o} m_{ij}+\tau}{N_{o}(1-m_{ij})+\tau}\right), \label{eq_g}
\end{equation}
where $\tau$ is an arbitrary additive constant (0.5 was suggested by Bartleson(1984) and is also been used in our experiment). Logistic matrix $\mathcal{G}$ are then transformed to $z$-scores matrix $\mathcal{Z}=(z_{ij})_{1\leqslant i,j\leqslant n}$ by using a simple scaling in the form of $\mathcal{Z}=\lambda \ast \mathcal{G}$ where coefficient $\lambda$ is determined by linear regression between z-scores and corresponding logistic values. We calculate this value of $\lambda$ to be 0.5441 with $N_{o}=3000$ observations for each pair of algorithms. From the matrix $\mathcal{Z}$, an accuracy score of composition method $i$ is calculated by averaging the scores of the $i$th column of the matrix $\mathcal{Z}$. The final matching scores are contained in an interval around 0. The interval can be estimated as $\pm 1.96/\sqrt{2N_{o}}$ at the $95\%$ confidence level. The resulting $z$-scores and confidence intervals for all outfits are shown in Figure ~\ref{fig:scoreHigtogram}. It is evident that our model performs superiorly to other methods and auto-encoder is the least suitable. The matching scores of five composition methods are -0.7123, -0.1546, 0.1227, 0.2655 and 0.4788 separately with $\pm 0.0253$ confidence interval. Figure ~\ref{fig:compositionCompared} depicts directly visual comparative results of representative composition outfits.  From the figure, we can see that our method get more satisfactory composition results. The method with score label gives more reasonable composition. It is obvious that our method produces better composition outfit than the other methods.

\begin{figure}
\centering
\includegraphics[scale =0.49]{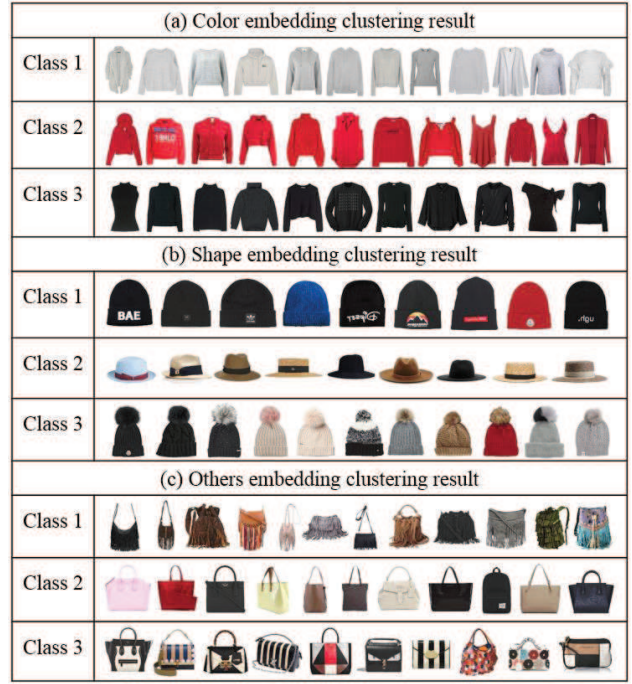}
\caption{items clustering visual result with different parts of embedding.}
\label{fig:colorShapeOther}
\end{figure}

\subsection{Validity of Attribute Embedding}

In our experiment, we adopt clustering method to validate the discriminative ability of the encoded attributes embedding. After finishing training of the whole network, we use encoding network to extract the embedding of items. Then, we use $k$-means methods to cluster attributes corresponding part of embedding. Figure ~\ref{fig:colorShapeOther} gives some random picked visual clustering results. Figure ~\ref{fig:colorShapeOther}(a) shows that tops in the same class are similar in color attributes, which proves that the extracted color corresponding parts of embeddings are distinguishable.
%
%
%
%
In Figure ~\ref{fig:colorShapeOther}(b), items that have exactly similar shapes are clustered into the same class. To verify the effectiveness of mutually independent constraint, we use remaining attributes corresponding parts of embeddings to cluster bag items. From Figure ~\ref{fig:colorShapeOther}(c), we can observe that bags in the same class usually have some attributes in common. Bags in class 1 and class 2 have similar texture and bags in class 3 have similar styles, which demonstrates that remain feature includes other useful attribute feature. Meanwhile, bags in the same class usually have multifarious shape and color, which proves that the mutually independent constraint works.

\subsection{Personalized Attributes Composition}

\begin{figure}
\centering
\includegraphics[scale =0.36]{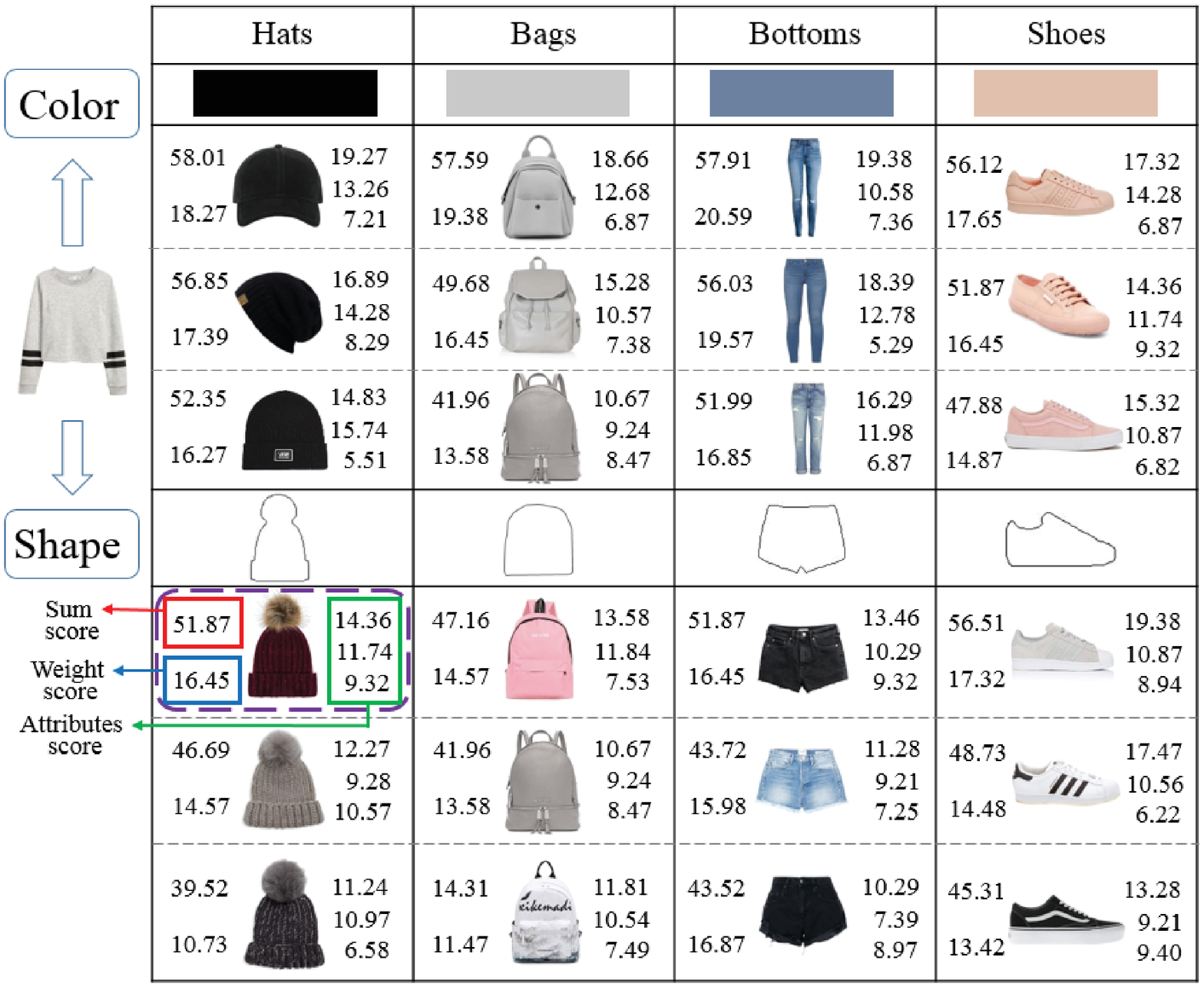}
\caption{Outfit Composition with user specify attributes (color or shape).}
\label{fig:compositionUserSpecified.}
\end{figure}

In the stage of building composition graph, we cluster the attributes corresponding parts of embeddings into different level cluster centers, which are set as vertexes in the graph. In the testing stage, users can specify their preferred attributes, such as ``color=red'' and ``shape=circle'', our model can give a series of specified attribute items with interpretable matching scores for users to choose. Figure ~\ref{fig:compositionCompared} shows some ordered recommended matching items with five composition score. For example, sum composition score of  the recommended hat in purple dashed box is 51.87. Weight composition score in blue box is 16.45. Attributes matching scores corresponding to color, shape and remaining attributes are 14.36, 11.74 and 9.32, respectively. From attributes matching scores, we can see that color attribute and shape attribute give more contribution to the final composition than remaining attributes, which is consistent with our questionnaire's result. For the specified color attribute, recommended items are almost indiscriminate in color. For the specified shape, all recommended items have similar shapes that are in line with the original specified shape.

\subsection{Outfit Composition's Extension}
%
%
%
%
As described above, our fashion composition graph is weakly supervised, which only dependents on outfit dataset without matching score labels. Our model can also be extended to a dataset with matching score labels. After normalizing the favorite user scores, the fashion composition graph with the score is built using Eq.(\ref{eq:w}) by setting $\alpha$ a normalized score of the corresponding outfit. Outfit composition result with score label are given in Figure ~\ref{fig:compositionCompared}, Figure ~\ref{fig:scoreHigtogram} and table 2.

\begin{figure}
\centering
\includegraphics[scale =0.30]{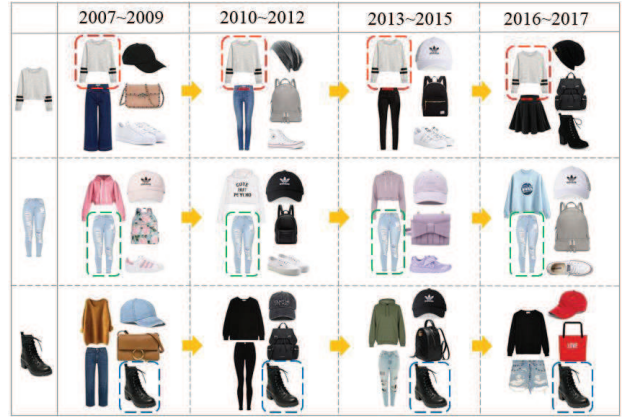}
\caption{Fashion trends with different years}
\label{fig:fashionTrend}
\end{figure}

The fashion trends is known by its quick and dogmatical variation ~\cite{abe2017changing}. The proposed technique enable us to keep up with the fashion trend along time. When new trend outfit dataset joints, old $W$ of composition graph is divided by $\sqrt{N_{m}}$, where $N_{m}$ is the max weight of the graph. Then, the fashion composition graph is updated following Eq.(\ref{eq:w}) with $\alpha>1$. In the experiments, we classify our whole outfit dataset into four parts according to years: dataset1(2007-2009), dataset2(2010-2012), dataset3(2013-2015), dataset4(2016-2017). We use our proposed iterated model with dataset1(2007-2009) as the basic dataset to get initial outfit composition graph. Then, by constantly adding new dataset into the basic dataset, we can get new outfit composition graph, by which fashion trend is kept up with. Figure ~\ref{fig:fashionTrend} gives the visual most popular outfit along years. We can see that our iterated model can keep up with fashion trends tightly.

%
%
%
%

\section{Conclusion}

 Fashion style tells a lot about one's personality. Interpretable embeddings of items are fatal in design, market and clothing retrieval. In this paper, we propose a partitioned embedding network, which can extract interpretable embedding from items. With attributes label as the constraint, different parts of embedding are restrained to related to corresponding attributes. With multi-independent constraint, different parts of embedding are restrained to only related to corresponding attributes. Then, using the extracted partitioned embeddings, a composition graph with attributes matching map are built. When users specify their preference attributes, such as color and shape, our model can recommend desirable outfit with interpretable attributes matching scores. Meanwhile, extensive experiments demonstrate that interpretable and partitioned embedding is helpful for designer, businesses and consumers to better understand composition in outfits. In applications, people's skin color and stature have great influence on outfit composition. Thus, personalization would be took into consideration in our future work. In further, straightforward composition relationship among items is another direction of our future work.

{\small
\bibliographystyle{ieee}
\bibliography{egbib}

}

\end{document}